# Evolving Label Usage within Generation Z when Self-Describing Sexual Orientation

Wilson Y. Lee[1]    J. Nicholas Hobbs[1]

1 The Trevor Project, West Hollywood, California
wilson.lee@thetrevorproject.org (WYL)

## Abstract

Evaluating change in ranked term importance in a growing corpus is a powerful tool for understanding changes in vocabulary usage. In this paper, we analyze a corpus of free-response answers where 33,993 LGBTQ Generation Z respondents from age 13 to 24 in the United States are asked to self-describe their sexual orientation. We observe that certain labels, such as bisexual, pansexual, and lesbian, remain equally important across age groups. The importance of other labels, such as homosexual, demisexual, and omnisexual, evolve across age groups. Although Generation Z is often stereotyped as homogenous, we observe noticeably different label usage when self-describing sexual orientation within it. We urge that interested parties must routinely survey the most important sexual orientation labels to their target audience and refresh their materials (such as demographic surveys) to reflect the constantly evolving LGBTQ community and create an inclusive environment.

## 1. Introduction

Sexual orientation can be defined as a person's self-identification of their sexual orientation, sexual behavior, sexual attraction, and emotional or romantic attractions (Park, 2016; Human Rights Campaign, 2022). In 2019, a survey asking about the sexual orientation of LGBTQ respondents found that 1 in 5 respondents identified as something other than lesbian, gay, or bisexual. (The Trevor Project, 2019). The set of vocabulary and specific labels reflecting the diversity of sexual orientation has continued to evolve beyond only the labels "lesbian", "gay", and "bisexual". Some sexual orientation labels have become more commonly used while others have declined. Famously, the meaning for the term *queer* has evolved its meaning across history and contexts as the younger generations of the LGBTQ community reclaimed the term from its derogatory connotations (Clarke, 2021).

Understanding the words that exist in a person's vocabulary and the importance these individual words hold in a person's vocabulary is a key aspect of understanding how a person expresses themself and how they interact with others. Failure to obtain such understanding often leads to unintentional harm. For example, the question "are you queer?" evokes vastly different feelings when directed at a 14-year-old vs a 60-year-old in 2022. The former is likely to answer the question with empowerment; the latter is likely to react with distaste. In this paper, we are interested in understanding the most recent generation to enter secondary schools and the workforce, Generation Z, and its vocabulary usage in regards to sexual orientation.

Generation Z is often defined as the generation cohort that is born between 1997 and 2012. It has the highest percentage of LGBT identifying individuals out of every generation at 20.1% (Dimock, 2019; Jones, 2021). Due to the larger percentage of individuals identifying as LGBT in Generation Z, it is more significant than ever to understand how this generation uses sexual orientation labels. Will Generation Z continue to evolve the usage of sexual orientation labels? Or will they begin to hold the usage of these labels constant? Little work has been done to determine the answer to these questions. In this paper, we examine the usage of sexual orientation labels from members of the LGBTQ community of Generation Z. We find highly consistent trends that inform us about the staying power of certain labels and demonstrate that even within this single generation, there still exists intra-generation differences in the usage of sexual orientation labels across individuals of different ages.

## 2. Method

### 2.1. Data Collection

We obtain data that was collected from an online nonprobability sample of 33,993 LGBTQ respondents ages 13 to 24 in the United States (The Trevor Project, 2022). Of the 33,993 LGBTQ respondents suveyed, 32,061 provide an answer to the question we analyze. The sample is recruited via targeted ads on social media between September 20 and December 31, 2021.

In order to understand the vocabulary LGBTQ respondents adopt to describe their own sexual orientation, we analyze responses to the open-response question *"Sexual orientation is a person's emotional, romantic, and/or sexual attractions to another person. There are many ways a person can describe their sexual orientation and many labels a person can use. How would you describe your current sexual orientation in your own words?"*.

## 2.2. Age Partitions

We apply *working demographic partitions* as defined by Lebart (1998) to allocate respondents into groups that are as homogeneous as possible. This task in our research is trivial as our sampling method (as described in section 2.1) asks respondents to self-report age and filters to exclude out-of-target samples.

We allocate all respondents into 12 distinct age partitions of interests with each corresponding directly to an age, for example group 1 consists exclusively of respondents of age 13, group 2 of age 14, etc.

| Partition | Size | Partition | Size |
|---|---|---|---|
| Age 13 | 3388 | Age 19 | 2194 |
| Age 14 | 4421 | Age 20 | 1787 |
| Age 15 | 4716 | Age 21 | 1742 |
| Age 16 | 4544 | Age 22 | 1499 |
| Age 17 | 4158 | Age 23 | 1478 |
| Age 18 | 2708 | Age 24 | 1358 |

Table 1. The sample size of documents for each age partition.

## 2.3. Measure Differences across Age Partitions

In order to measure how term importance changes as age increases, we start with $Corpus_{age<=13}$ and iteratively grow the corpus by joining the corpus with the next higher age partition to form $Corpus_{age<=14}$, $Corpus_{age<=15}$, etc.

Instead of measuring term importance unique in each age partition (i.e. $Corpus_{age=13}$, $Corpus_{age=14}$, etc), this corpus growth approach is more resistant to sample size differences across the age partitions and provides more robust comparability across corpora as each $Corpus_{age<=n}$ also includes all documents from $Corpus_{age<=n-1}$.

## 2.4. Data Cleaning and Transformation

We first perform pre-tokenization in order to parse the raw text into individual tokens suited for analyses (Mielke et al, 2021). We adopt whitespace pre-tokenization as implemented by Hugging Face that split sentences on word boundaries, including whitespace and non-Unicode characters, such as punctuations (Thomas et al, 2020). For example, "I am attracted to women!" would be pre-tokenized into the following list of tokens: "I", "am", "attracted", "to", "women", "!".

To ensure focus on sexual orientation labels, we filter out all punctuations, numerics and common English stopwords as defined by the Natural Language Toolkit (NLTK) from the list of tokens to be analyzed (Bird et al, 2009).

For each of the demographic partitions described in Section 2.2, we form a corresponding individual corpus where each document in the corpus is a list of cleaned tokens computed from the raw text of an individual survey response.

## 2.5. Ranked Term Importance Calculation

We apply term frequency–inverse document frequency (TF-IDF) to measure term importance. TF-IDF is a common statistical measure used to capture the relative importance of terms in a document by weighing raw term frequency with term specificity, calculated as inverse document frequency (Manning et al, 2010).

In order to provide focus in our analyses and derive insight from terms of the highest impact, we choose a threshold to only consider the top 100 most important terms in each corpus (Lebart, 1998).

At every iterative step described in Section 2.3, we calculate 1-gram TF-IDF on the entire new corpus. Comparing the TF-IDF results allows us to measure the change in importance rank for individual terms as the corpus grows to include responses from older respondents. We measure the rate of change for the TF-IDF rank with unit ($\frac{\Delta rank}{\Delta age}$).

## 3. Results

### 3.1. Consistent Importance

The consistent importance of sexual orientation labels was determined by calculating which labels did not have their ranks change across all age partitions. These labels include: bisexual, pansexul, and asexual. These labels were also the first, second, and third most important labels respectively for all the age partitions.

Additionally, there were sexual orientation labels with nearly consistent rank across all age partitions. These labels only shifted by one rank when observed through all age partitions. These labels include: asexual and gay. These labels were the fourth and fifth most important labels, depending on the age partition being examined.

The most important labels were also the most consistent words throughout the analyses. They had the least amount of change while maintaining the highest importance throughout the age partitions.

### 3.2. Evolving Importance

A number of sexual orientation labels also showed a strong evolution of importance based on the age partitions. These were determined by finding the largest magnitude rates of change and examining these labels. The list of labels included: homosexual (-2.356), demisexual (-1.430), omnisexaul (1.094), ace (-0.842), demiromantic (-0.517), and queer (-0.506). Values in parenthesis show the rate of change in ranked importance over age partitions, with **negative values showing an increase in importance** for increased age partitions, and **positive values showing a decrease in importance** for increased age partitions.

These results demonstrate that the sexual orientation labels homosexual, demisexual, ace, demiromatnic, and queer gain importance as responses from older respondents were included. On the other hand, the label omnisexual demonstrates the opposite trend. It should also be called out that the rates of change are not unidirectional. The label demiromantic in Figure 1 shows a mix of positive and negative rates of change.

| Label | Rate of Change ($\frac{\Delta rank}{\Delta age}$) | Avg. Rank |
|---|---|---|
| bisexual | 0.000 | 1.000 |
| pansexual | 0.000 | 2.000 |
| lesbian | 0.000 | 3.000 |
| asexual | 0.112 | 4.667 |
| gay | -0.112 | 4.333 |

Table 2. The calculated rate of change and average rank for evolving importance terms of constant or nearly constant labels.

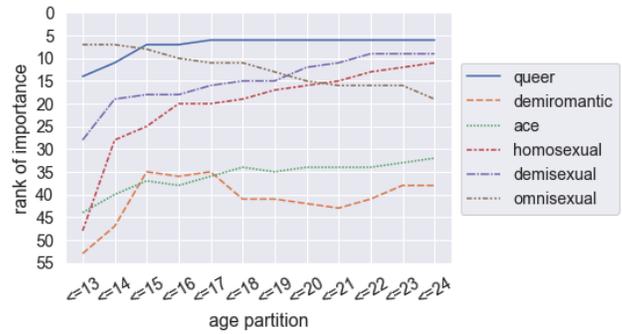

Figure 1. Rank importance over age partitions for the top 6 labels with the most change in rank importance. Lower number in rank means higher importance, i.e. 1 is the most important

| Label | Rate of Change ($\frac{\Delta rank}{\Delta age}$) | Avg. Rank |
|---|---|---|
| homosexual | -2.356 | 20.333 |
| demisexual | -1.430 | 14.917 |
| omnisexual | 1.094 | 12.417 |
| ace | -0.842 | 35.917 |
| demiromantic | -0.517 | 40.833 |
| queer | -0.506 | 7.250 |

Table 3. The calculated rate of change and average rank for evolving importance terms of magnitude 0.50 and greater. Negative rate of change means more importance for older age partitions.

## 4. Discussion

### 4.1. Intra-subgroup Differences within Generation Z

Although generation cohorts (especially Generation Z) are often stereotyped to behave homogeneously in mass media, theory of generations and generation units states that subgroups exist within a generation cohort, for example, along political, class or other lines (Mannheim, 1952). Other works theorize that although shared experience as queer individuals do form such a subgroup within a generation cohort, there is natural tension that arises from insisting on a homogeneous narrative across all LGBTQ experiences (Marshall, 2019). The results in this paper extend on exploring this tension and provide evidence that even within a generation cohort of LGBTQ individuals, we still observe noticeably different vocabulary adoption in one of the key identities that define membership in the LGBTQ subgroup (i.e. sexual orientation).

Unlike Mannheim, our work stops at exploring other social factors, such as class differences, and takes on an intentionally simplistic view in forming subgroups based on discrete age. This design allows us to linearly grow the corpus by age at each iterative step and directly observes how *difference in age* impacts term importance.

### 4.2. "Homosexual": the sharpest decline in importance among younger individuals

Out of all examined labels, the term importance of the label "homosexual" shows the highest magnitude of change across age partitions, with responses from younger individuals exhibiting less importance for the term. In 2014, The New York Times published an article titled "The Decline and Fall of the 'H' Word" that suggests that many scholars expect the use of the term "homosexual" to eventually fall away entirely and to be substituted with terms "gay" and "lesbian" (Peters, 2014). Although our data does not provide a causal explanation for this phenomenon, our results of the terms "gay" and "lesbian" showing consistenly high importance across all age partitions also provide further evidence that points to this trend.

The 2014 New York Times article also asserts that a scan for the term "homosexual" on the Google Books Ngram Viewer shows a consistent decline of the term in recent years after peaking in popularity around 1995 (Michel et al, 2011). As of 2022, GLAAD, a US-based non-profit dedicated to LGBT advocacy in media, recommends on its webiste "[a]void identifying gay people as 'homosexuals' an outdated term considered derogatory and offensive to many lesbian and gay people" (GLAAD, 2022). As more commanding sources in the LGBTQ community continues to make similar recommendations, we hypothesize that we will continue to observe this trend year over year until the label "homosexual" fully falls out of the top most important sexual orientation labels in a similar study.

### 4.3. "Omnisexual" and "Demisexual"

Among the examined labels, only one label "omnisexual" exhibits an unidirectional positive rate of change. In other words, "omnisexual" is the only label where its importance declines as samples from older individuals are added to the term importance analysis. The label "demisexual" exhibits a similar trend in the opposite direction, as the label's importance increases as the sample gets older. Aside from the opposite in trend direction, "omnisexual" and "demisexual" show similar absolute magnitude in rate of change and similar average rank across all age partitions. We do not have the evidence in our data or sufficient theory in literature to offer hypotheses on what causes this phenomenon, but we point out these observations in this paper in hope to inspire other research, especially surveys tailored to understanding these labels.

### 4.4. Labels with Consistent Importance

In previous sections, we have highlighted and provided hypotheses on labels that show evolving importance across age partitions. It is also worth noting that there are labels that show consistent importance among all age partitions. It is perhaps not surprising that "lesbian", "gay" and "bisexual" show consistent importance as the three labels are prominently displayed in the popular acronym LGBT. We hypothesize that the relatively higher visibility and familiarity from mentions of the acronym in media leads to the labels' consistent importance. These are also the three labels that often appear in government documents, for example the 2020 US Census (Anderson et al, 2021).

Furthermore, the LGBT+ Pride 2021 Global Survey conducted by IPSOS shows that Generation Z is significantly more likely to identify as gay, lesbian, bisexual, asexual and pansexual compared to older generations. Our results showing consistent and highly ranked importance for all five of these labels across all age partitions provide additional evidence for this observation.

### 4.5. Limitations

It is important to note that there are notable limitations to this methodological setup. Our naive approach to pre-tokenization of the raw text does not take into account any context. "I am gay" and "I am not gay" would contribute equally to the term importance calculation. Phrases such as "omnisexual or pansexual" would contribute to importance calculation for both of the labels "omnisexual" and "pansexual". Furthermore, we do not attempt to interpret the meaning of the text to form associations between responses containing similar semantic meaning but vastly different terms, for instance "questioning" vs "idk" and "I am attracted to men" vs "homosexual".

Through these examples, we stress that there is an important distinction between held sexual orientation identity and the vocabulary a respondant adopts to describe it. In this study, we are only studying the latter–the vocabulary. All of the conclusions we draw are about the trends on the *words* respondents choose to use to describe their identities and not necessarily about the trends on the held identities. For instance, the decreasing importance of the label "homosexual" in younger respondents does *not* suggest younger respondents hold less importance in the homosexual identiy, but rather they hold less importance in using the word "homosexual" to describe their identity.

Lastly, as previously stated, Generation Z is often defined as the generation cohort that is born between 1997 and 2012 (Dimock, 2019). At the time of this survey in 2021, Generation Z roughly translates to individuals with age ranging from 9 to 24. Our sample does not include any respondents from age 9 to 12; 4 Generation Z age partitions are therefore missing in our analyses. Although it is incomplete, we believe that the 12 age partitions (from age 13 to 24) still present meaningful insights into Generation Z.

## 5. Conclusion

We evaluate the ranked importance of sexual orientation labels from corpora consisting of open responses to self-describe sexual orientation and observe changes across age partitions within members of the LGBTQ Generation Z community. Our results show that members of the LGBTQ Generation Z community are not homogenous and adopt different vocabularies when describing their sexual orientation based on their age. Key findings include "bisexual", "pansexual", "lesbian", "asexual", and "gay" having consistent importance regardless of age within the generation. In contrast, "homosexual", "demisexual", "ace", "demiromantic", and "queer" show a changing level of importance over age partitions.

These findings provide evidence in support of known trends, such as the recent decline in the use of the word "homosexual".  While also showing evidence for previously unobserved trends, such as "demisexual" having more importance for older respondents and "omnisexual" having more importance for younger respondents. We encourage other researchers to conduct further studies to investigate these observations.

To date, there has been little large-scale quantitative research into sexual orientation labels as lexicons that evolve importance over time and hold different semantic meanings across generations and within a single generation. We urge researchers to not only conduct studies to understand the held identities but also the words members of the LGBTQ community adopt to describe them. Future directions include examining trends in label term importance within Generation Z along other sub-group social factors, such as race, geographic region, gender identity, etc, and comparing Generation Z to other generation cohorts, such as Millennial and Generation X.

Reflecting on the practical social impact of understanding sexual orientation lables and their relative importance, we echo the assertions made by DeChants in 2021 that "[n]eglecting to collect information about LGBTQ identities contributes to stigma" and "[a] failure to capture LGBTQ identities in public health data collection can further perpetuate existing inequities". For materials where inclusive sexual orientation labels should be expected, the lack of labels contributes to stigma for individuals and erases the identities of those individuals. The findings presented here can help these materials become more inclusive by encouraging others to understand the continual shifting usage of sexual orientation labels even within a single generation cohort. We encourage other researchers to build on our findings to help reduce stigma and inequities against the LGBTQ community.

[12] Mannheim, K. (1952). The problem of generations. In D. Kecskemeti (Ed.), Essays on the sociology of knowledge (pp. 276–322). Routledge & Kegan Paul.

[13] Manning, C.D.; Raghavan, P.; Schutze, H. (2008). "Scoring, term weighting, and the vector space model". Introduction to Information Retrieval. p. 120.

[14] Marshall, D., Aggleton, P., Cover, R., Rasmussen, M. L., & Hegarty, B. (2019). Queer generations: Theorizing a concept. International Journal of Cultural Studies.

[15] Michel, J. B., Shen, Y. K., Aiden, A. P., Veres, A., Gray, M. K., Google Books Team, ... & Aiden, E. L. (2011). Quantitative analysis of culture using millions of digitized books. science, 331(6014), 176-182.

[16] Mielke, S.J., Alyafeai, Z., Salesky, E., Raffel, C., Dey, M., Gallé, M., Raja, A., Si, C., Lee, W.Y., Sagot, B. & Tan, S., (2021). Between words and characters: A Brief History of Open-Vocabulary Modeling and Tokenization in NLP. arXiv preprint arXiv:2112.10508.

[17] Park, A. (2016). Reachable: Data collection methods for sexual orientation and gender identity (p. 4). The Williams Institute.

[18] Peters, J.W. (2014, March 21). *The Decline and Fall of the 'h' Word.* The New York Times. https://www.nytimes.com/2014/03/23/fashion/gays-lesbians-the-term-homosexual.html

[19] The Trevor Project. (2019). *Diversity of Youth Sexual Orientation.* West Hollywood, California: The Trevor Project. https://www.thetrevorproject.org/research-briefs/diversity-of-youth-sexual-orientation/

[20] The Trevor Project. (2022). *2022 National Survey on LGBTQ Youth Mental Health.* West Hollywood, California: The Trevor Project. https://www.thetrevorproject.org/survey-2022/assets/static/trevor01_2022survey_final.pdf

[21] Wolf, T., Debut, L., Sanh, V., Chaumond, J., Delangue, C., Moi, A., ... & Rush, A. M. (2020, October). Transformers: State-of-the-art natural language processing. In Proceedings of the 2020 conference on empirical methods in natural language processing: system demonstrations (pp. 38-45).